\documentclass{article}
\usepackage{spconf,amsmath,epsfig}
\usepackage{tikz}
\usepackage{xcolor}
\usepackage{color}
\usepackage{graphicx}
\usepackage{algorithm2e}
\usepackage{breqn}
\usepackage{subfigure}
\usepackage{url}
\usepackage{booktabs}
\usepackage{enumitem}

\usepackage{etoolbox}
\patchcmd{\thebibliography}
  {\settowidth}
  {\setlength{\itemsep}{0pt plus 0pt}\settowidth}
  {}{}
\apptocmd{\thebibliography}
  {\small}
  {}{}
  
\newcommand{\ie}{i.e.}

\newcommand{\m}[1]{{\mbox{{\fontencoding{T1}\sffamily\slshape{#1\/}}}}}

\title{Ocean Eddy Identification and Tracking using Neural Networks}
\name{Katharina Franz, Ribana Roscher, Andres Milioto, Susanne Wenzel, J\"urgen Kusche}
\address{University of Bonn, Institute of Geodesy and Geoinformation, Nussallee 15+17, 53115 Bonn, Germany\\
}
\begin{document}
%
\maketitle
\begin{abstract}
Global climate change plays an essential role in our daily life. 
Mesoscale ocean eddies have a significant impact on global warming, since they affect the ocean dynamics, the energy as well as the mass transports of ocean circulation. 
From satellite altimetry we can derive high-resolution, global maps containing ocean signals with dominating coherent eddy structures.
The aim of this study is the development and evaluation of a deep-learning based approach for the analysis of eddies. 
In detail, we develop an eddy identification and tracking framework with two different approaches that are mainly based on feature learning with convolutional neural networks.
Furthermore, state-of-the-art image processing tools and object tracking methods are used to support the eddy tracking.
In contrast to previous methods, our framework is able to learn a representation of the data in which eddies can be detected and tracked in more objective and robust way. 
We show the detection and tracking results on sea level anomalies (SLA) data from the area of Australia and the East Australia current, and compare our two eddy detection and tracking approaches to identify the most robust and objective method. 
\end{abstract}
\vspace{-0.3cm} 

\section{Introduction}
\label{sec:intro}
\vspace{-0.05cm} 
The global ocean circulation is substantially affected by characteristic vortices, called mesoscale ocean eddies, moving circularly in parts of the Earth's great currents.
Their main contribution to ocean circulation is due to the transport of a huge amount of kinetic energy. 
Hence, their detection and tracking is highly relevant in terms of oceanography and climate change studies, respectively.
Pattern recognition in satellite imageries is a common task in remote sensing, where deep learning methods make the large amounts of data collected from space considerably more usable than traditional machine learning algorithms. 
These approaches allow the recognition of objects and coherent patterns from data, wherefore we develop a convolutional neural network (CNN)-based eddy detection and tracking framework.

Mesoscale eddies are small circular currents that show characteristic cyclonal and anticyclonal patterns in global maps, e.g. sea level heights~\cite{chelton:2010}, and their automatic detection and identification is an active research topic. 
Several methods have been developed to approach the formulated detection task, some of which are geometrically inspired~\cite{sadarjoen:2000}, or on the basis of signal decompositions~\cite{doglioli:2007}. In particular, the Okubo-Weiss method~\cite{Okubo:1970,Weiss:1991} is the most popular physical procedure to approach this problem, due to its physical interpretability, but open questions remain with respect to the threshold definiton~\cite{Faghmous:2014}.
On the other hand, deep learning algorithms are a breakthrough innovation to approach computer vision tasks such as object recognition, and have recently been used for oceanographic applications (\cite{eddynet:2017}). 
However, neural networks require a large amount of reference data, usually only available at a high cost, in order to learn an adequate model for recognition.
Therefore, these supervised approaches are combined with data generation or data augmentation procedures to overcome this problem.
\cite{ValletS15,pathak:2017}, for example, propose automated techniques to generate training data. In contrast \cite{Yosinski:14} suggest transfer learning approaches based on pretrained networks on large openly available datasets that are fine-tuned for the current classification task which significantly reduces the amount of training data needed.
\cite{kemker:2017} develop a framework for the classification of multispectral images using deep learning techniques by segmentation with CNNs, and consider the scarcity of annotated satellite data by proposing the synthesization of reference data as an automated procedure.

In this paper, we propose an eddy detection and tracking framework combining feature learning by CNNs with an established image processing tool, the Kanade-Lucas-Tomasi (KLT) ~\cite{lucasKanade81:klt} feature tracker. Furthermore, we compare this with a recurrent neural network (RNN), in particular a long-short term memory (LSTM), trained for eddy identification.
We utilize the Okubo-Weiss method to tackle the lack of annotated data by detecting a few, yet precise eddies, which are used as training data.
By combining these approaches, we are able to achieve a high recall as well as high precision, which tradeoff is generally a problem for ordinary eddy detection method such as the Okubo-Weiss procedure.\vspace{-0.3cm} 

\section{Data}
\label{sec:data}
\vspace{-0.05cm} 
Our study site covers seas around Australia and the South Equatorial current.
The dataset contains high-resolution, global sea level anomaly maps (SLA), which are preprocessed products from the ESA Climate Change Initiative (ESACCI)~\cite{ESA:2015}.
The gridded SLA maps contain the mesoscale coherent eddy features and variabilities and enable the distinction between cyclonal and anticyclonal eddies. 
The data set covers daily data from January 1993 until December 2014, merged from several altimetry missions.
The training and validation data sets are generated automatically (cf. Subsec.~\ref{ssec:pgt}), and state independent regional patches from the global SLA maps with a length of 365 days per year.
The validation data set is a smaller subset of the training set.
We represent the data as 1-band images with a spatial resolution of $0.25^\circ$ in a $256 \times 256$ and in a $49 \times 49$ grid. 

\begin{figure}[tb]
\centering
\includegraphics[width=0.38\textwidth]{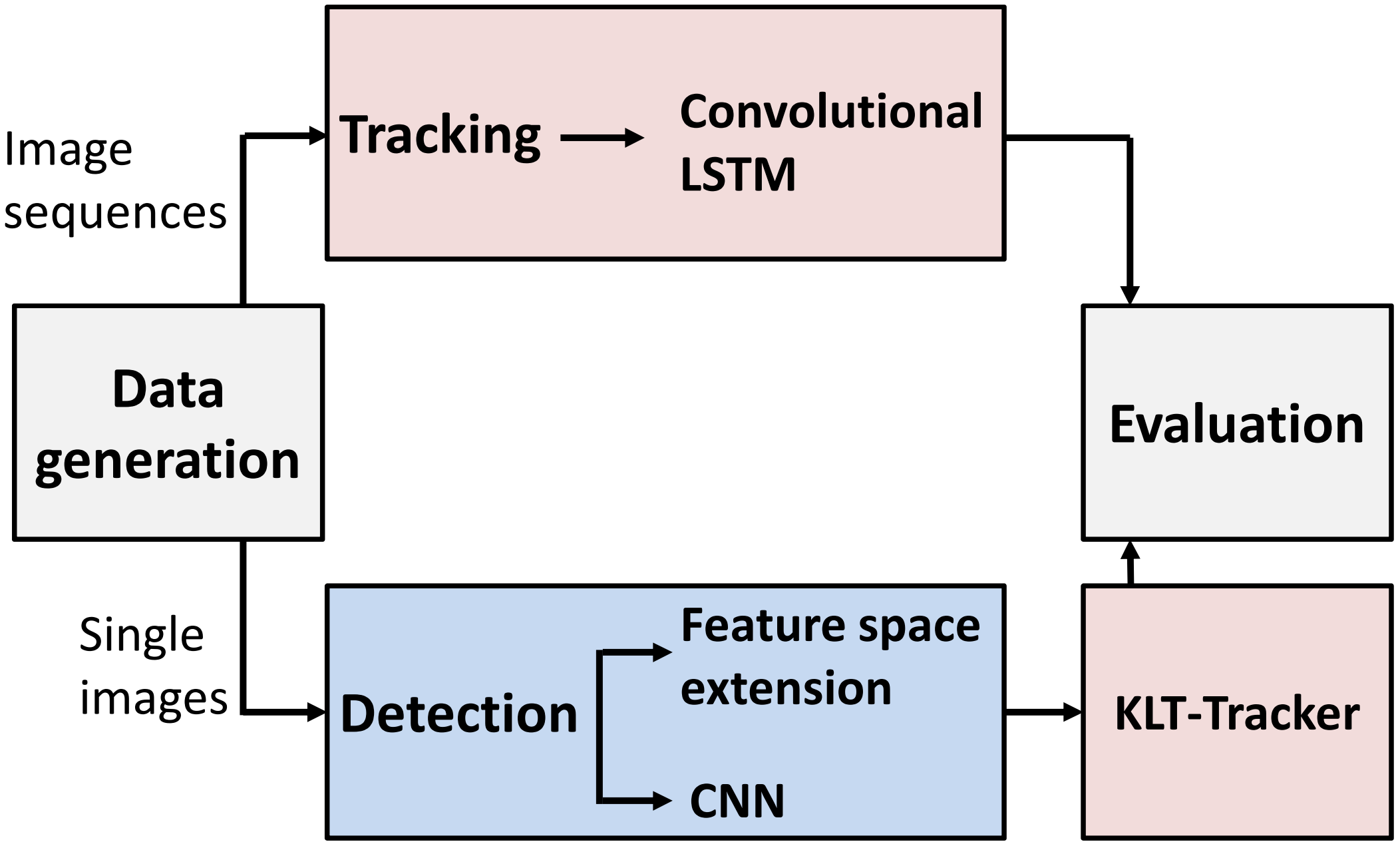}
\vspace{-0.4cm}
\caption{{\bf Workflow:} Annotated data and one-channel SLA images are generated serving as input to the detection module and the tracking module. Two approaches, a CNN+KLT-Tracker and a convolutional LSTM are applied and evaluated.
\vspace{-0.3cm}}
\label{fig:workflow}
\vspace{-0.3cm}
\end{figure}

\section{Method}
\label{sec:method}
\vspace{-0.05cm} 
Our framework combines CNNs for feature learning with two different tracking approaches, namely a tracking realized on the basis of an optical flow estimation with KLT-Tracker, and an RNN. 
It contains different modules that are visualized in a flowchart in Figure ~\ref{fig:workflow}.
\begin{enumerate}[leftmargin=0.5cm]
\setlength\itemsep{0em}
	\item \textbf{Data generation:} The pseudo-reference data module generates annotated data, which we split into training, validation and test sets. Corrected eddy detections from Okubo-Weiss method serve as basis for this module, see Sec.~\ref{ssec:pgt}.
	\item \textbf{Detection:} The CNN-module learns a model from which we can detect eddy cores at a single epoch, see Sec.~\ref{ssec:deepArch}.
	\item \textbf{Tracking:} The KLT-tracker tracks the detected eddies using a sparse optical flow. On the other hand, a RNN uses convolutional LSTM units to track the eddy cores, see \ref{ssec:imageProc}.
    \item \textbf{Evaluation:} Since no reference data is given, we evaluated our results qualitatively with focus on plausability.
\end{enumerate}

\vspace{-0.2cm} 
\subsection{Pseudo-reference data generation with Okubo-Weiss method}
\label{ssec:pgt}
\vspace{-0.05cm} 
In order to tune and train the neural network, we require annotated reference data, which generally does not exist in our application. 
Consequently, we propose an automated labeling process to generate precise pseudo-reference data based on the results of the Okubo-Weiss method~\cite{Okubo:1970,Weiss:1991,Cheng:2014}.


The Okubo-Weiss method is an established algorithm for eddy detection.
It aims at the computation of a physically motivated parameter that represents the balance between vorticity scales and shear strain rates. 
The value of Okubo-Weiss-parameter decides whether the vorticity dominates over strain, where an eddy core is defined as vorticity dominated area.
Therefore, the choice of the threshold on the Okubo-Weiss-parameter defines the precision and recall of the eddy identification.
Based on this, we define a threshold which provides a few, yet precise eddy cores, which are corrected manually and used as training and validation data for the CNN.
We treat the task as a two-class problem such that each cell of the grid-based data is assigned to the class 'eddy' or 'non-eddy', nevertheless, the approach can be extended to distinguish between anticyclonic, cyclonic and non-eddies.
Since each eddy core can only be defined with a certain precision, we define a circular domain around each detection and treat it as eddy core rather than defining only one grid cell.

\vspace{-0.05cm} 
\subsection{Eddy identification with CNNs}
\begin{figure}[tb]
\centering
\includegraphics[width=0.5\textwidth]{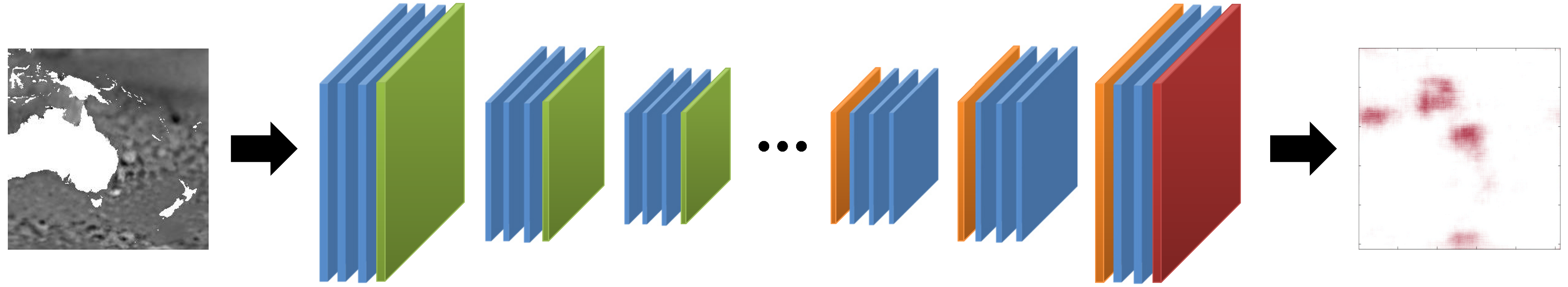}
\vspace{-0.8cm}
\caption{{\bf CNN architecture:} Blue: Conv, BatchNorm and ReLU-activation layers; Green: pooling layers; Orange: upsampling layers; Red: softmax-activation layer.}
\label{fig:arch}
\end{figure}

\vspace{-0.05cm} 
Figure~\ref{fig:arch} visualizes the architecture of our CNN, which we realize as encoder-decoder. 
We use five CNN building blocks with convolutional, batch normalization and activation layers, and a pooling layer, respectively, in the encoder part. 
The decoder part is defined in a similar way as the encoder, except for the downsampling layer, for which we use an upsampling layer at the beginning of each building block.
Our input are gridded one-band image subsets. 
The output of the CNN are the probabilities of the classes 'eddy' or 'non-eddy', where the output has the same dimension as the input image due to the chosen encoder-decoder architecture.
In order to localize eddy cores, 
we extract local maxima using non-maxima suppression, and define these points as eddy cores. 
Furthermore, the eddy core identification can be optimized and improved in terms of robustness by extending the feature space, \ie, with velocity information.
\label{ssec:deepArch}

\vspace{-0.05cm} 
\subsection{Eddy tracking}
\label{ssec:imageProc}
\vspace{-0.05cm} 
In order to track identified eddies over time, we developed two approaches. 
First, we utilize the KLT-tracker with sparse optical flow estimations and, besides, we apply a convolutional LSTM network.
Optical flow is defined as distribution of apparent velocities and motion patterns, and therefore leads to spatial information and the changes of spatial arrangements of objects. 
The KLT-tracker is a feature tracker that traces distinctive points over time based on the motion field of optical flow estimations~\cite{lucasKanade81:klt}. 
Thus, the differences in motion between two consecutive time steps are defined as $\m I(x,y,t) = \m I(x + \Delta x, y + \Delta y, t + \Delta t)$,
where $\m I$ is the gridded SLA data with position $(x,y)$ at time $t$ and the corresponding differentials $\Delta x$, $\Delta y$ and $\Delta t$.
In our study, the positions are estimated by the detection method.
As second approach, we use RNNs which directly process time-varying image sequences of SLA data. In detail, we apply convolutional LSTMs, which are able to model spatiotemporal dependencies. \vspace{-0.3cm} 

\vspace{-0.05cm} 
\section{Experiments and Discussion}
\label{sec:experiments}
\vspace{-0.05cm}     
\subsection{Experimental Setup}
\vspace{-0.05cm} 
We investigate the CNN-based detections regarding robustness and amount, and their suitability as input into a KLT-tracker with sparse optical flow estimation. 
In contrast, we address the stand-alone tracking approach with convolutional LSTMs.
For our study, the threshold for the Okubo-Weiss method is set to $0.1\sigma_W$. Here, $\sigma_W$ is the standard deviation of the Okubo-Weiss parameter.
The radius of the eddy cores used as pseudo-reference data is defined as $5$ grid cells in each direction which corresponds to $135$~km.
For training the CNN, we use a kernel size of $3$ grid cells which is equal to the stride.   
Furthermore, we have trained the CNN on $100$ epochs and used a batch size of $18$ grid cells. However, the convolutional LSTM is trained on $300$ epochs since the net needs to fit more hyperparameters. For those computations, we take advantage of a GPU's power, although, we need to resize the images to $49 \times 49$ grids.
We extract training data from all time steps, except March $1993$, which we use for testing.
We use the data from March $1st$ 1993 for eddy detection, and the next four days for tracking.
The implementations for our framework are realized in MATLAB and Python. 
The Python interfaces are based on Keras with TensorFlow backend.

\vspace{-0.05cm} 
\subsection{Results}
\begin{figure*}[tpb]
  \centering
 \includegraphics[width=0.95\textwidth,trim={0 0 0 2cm},clip]{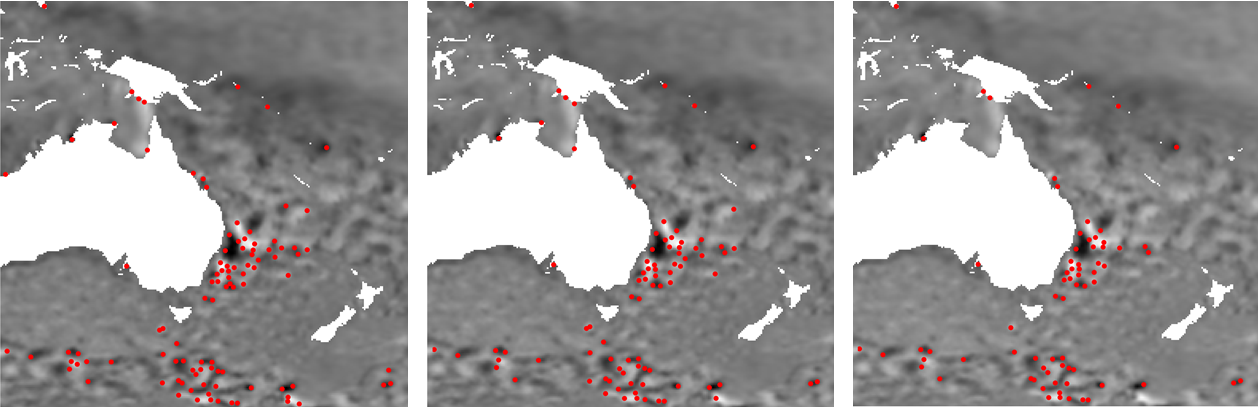}
\caption{{\bf: Eddy detections with CNN and tracking results obtained by KLT-tracker:} from left to right: March $1st$, March $2nd$ and March $3rd$ 1993; \textbf{red dots}: eddy cores)}
\label{fig:res}
\end{figure*}

\vspace{-0.05cm} 
Figure~\ref{fig:res} shows the results at the East Australian and the South Equatorial currents for each point in time. 
The left frame of the sequence presents the detected eddy cores at March $1st$ 1993, where eddy detections are colored in red.
The CNN-based detections and the tracked eddy cores with the KLT-tracker are visualized in Fig. \ref{fig:res}(a), in Fig. \ref{fig:res}(b) we show the eddy probability scores returned by the convolutional LSTM.\\
Eddies have a high variability, and therefore, they can appear, move or vanish between two consecutive time steps. Consequently, the number of detected eddies varies in the sequences, permanently. We collect the numbers of identifications in Table \ref{tab:cnn}.

\vspace{-0.05cm} 
\begin{table}[htb]
\centering
\caption{\textbf{CNN detections and KLT-tracking:} number of detected and tracked eddy cores (first row), averaged distances of eddy movement (second row)}
\label{tab:cnn}
\begin{tabular}{ l l l l l l }
\hline
& Day $1$ & Day $2$ & Day $3$ & Day $4$ & Day $5$ \\ \hline\hline
$\#$ Eddies & $100$ & $87$ & $71$ & $64$ & $62$ \\ \hline
Av. distance & - & $23~km$ & $20~km$ & $24~km$ & $25~km$ \\ \hline\vspace{-0.3cm} 
\end{tabular}
\end{table}

\vspace{-0.05cm} 
\subsection{Eddy Identification}\label{ssec:det}
\vspace{-0.05cm}  
From the CNN output, we extract most likely cells (see Sec. \ref{ssec:deepArch}). These detected eddy positions are colored in red  showing probabilities of about $60~\%$ and more. Overall, we detect $100$ eddies at March $1st$.
The CNN returns different intensities, \ie, a weaker eddy signal with low probabilities of about $62~\%$ is to be found in the South of Australia, despite this, we are able to identify the relevant eddy structures. The greatest concentration of eddy cores is detected at the warm East Australian Current. By comparison, the probabilities of $69~\%$ are slightly higher than those at the South of Australia. Weak eddy signals indicate high variabilities. In consequence, the most eddies might vanish in those fields and they may provide no sufficient tracking results. Furthermore, the CNN detects the highest probabilities near Papua-New Guinea in the North, and in the Northeast of Australia near the coast. 
Here, we extract $8$ eddy cores, which are located in the high-probability areas, as can be seen in Fig.~\ref{fig:res}(a). Nevertheless, each identified eddy core is characterized by an uncertainty, due to the noisy pseudo-reference data and the filtering algorithm applied on the CNN probabilities.

\vspace{-0.05cm} 
\subsection{Eddy Tracking}\label{ssec:tracking}
\vspace{-0.05cm} 
The results of the KLT-Tracker are presented in the SLA maps of day $2$ up to day $5$ in Fig. \ref{fig:res}. Tab. \ref{tab:cnn} counts the number of tracked eddy cores. 
Over time the number of tracks is reduced, hence, some eddies vanish in regions with high variability. These eddies are located at the East Australian Current and the Antarctic Circumpolar Current, in particular. 
Although, there is no apparent movement across the grid cells of the tracked eddies in Fig. \ref{fig:res}, we can analyze the movement of the eddies using the subpixel estimation of the KLT-tracker.
We computed the Euclidean distances between assigned points, and report the averages in Tab. \ref{tab:cnn}. 
A drawback of the KLT-tracker is a potential mismatch of tracked eddies due to their similar pattern, since all detections are tracked independently. 
In addition, the uncertainties of the estimations plays a role in imprecisions along the time series, too.
\begin{figure}[htpb]
  \centering
 \includegraphics[width=0.5\textwidth]{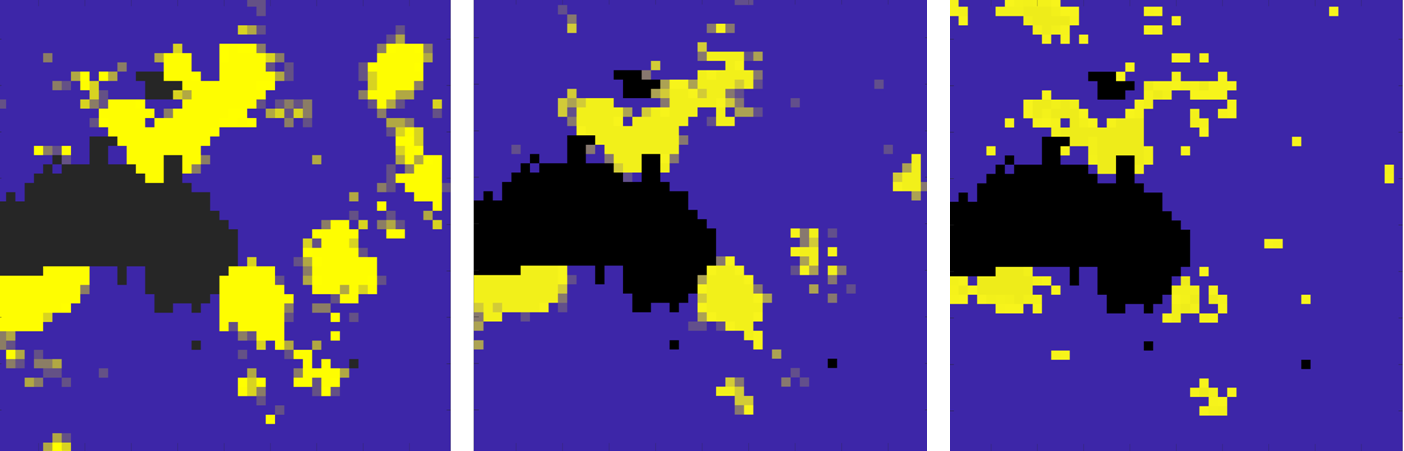}
\caption{{\bf: Eddy detections and tracking results using a convolutional LSTM:} from left to right: March $1st$, March $2nd$ and March $3rd$ 1993; \textbf{yellow:} high probabilities; \textbf{blue}: low probabilities; \textbf{black}: land area}
\label{fig:res2}
\end{figure}

Fig. \ref{fig:res2} shows our preliminary results of using a convolutional LSTM for eddy tracking. 
The results indicate that areas with multiple eddies can be detected. 
Due to memory and capacity restrictions of our used GPU, the spatial resolution is lower in comparison to our previous results, making a detection of single eddies difficult. 
However, we see this approach as a promising direction for future research, since it is capable to learn an appropriate eddy representation and the distribution of the eddy populations in order to track intense eddy signals.
\vspace{-0.2cm}

\section{Conclusions and Future work}\label{ssec:concl}
\vspace{-0.05cm} 
The convolutional LSTM reveals plausible and robust preliminary results. Hence, it is preferable to develop a convolutional LSTM that is able to distinct between single cyclonal and anticyclonal eddies. 
Furthermore, in order to make eddy detections more robust and precise, we propose to extend the feature space with dense optical flow information at each point. 
Further data like vorticity or sea surface temperatures are other options provided they are independent from SLA data. However, the spatial resolution is not adequate, such that the analysis of along-track data in place of gridded maps will be considered in the future .
{
\vspace{-0.05cm}
\bibliographystyle{IEEEbib}
\footnotesize
\bibliography{references.bib}
}
\end{document}